\documentclass[letterpaper, 10 pt, conference]{ieeeconf}  

\IEEEoverridecommandlockouts                              

\usepackage{graphicx}
\usepackage{hyperref}

\title{\LARGE \bf
PEACE: A Planner–Executor Agent with Constraint Enforcement for UAVs
}

\author{Erdem Uysal$^{1}$ and Timo Kehrer$^{1}$ and Sebastiano Panichella$^{1, 2}$%
\thanks{$^{1}$Institute of Computer Science,
        University of Bern, 3012 Bern, Switzerland
        {\tt\small \{ramazan.uysal, timo.kehrer, sebastiano.panichella\}@unibe.ch}}%
\thanks{$^{2}$AI4I - The Italian Institute of Artificial
        Intelligence, 10129 Torino, Italy 
        {\tt\small sebastiano.panichella@ai4i.it}}%
}

\begin{document}

\maketitle
\thispagestyle{empty}
\pagestyle{empty}

\begin{abstract}
Foundation models are increasingly used to drive autonomous
systems, yet existing approaches either keep the model in a tight 
control loop, raising latency and hallucination risk, or compile 
natural language into opaque end-to-end policies that are hard to 
explain, constraint and require domain-specific datasets and fine-
tuning. We propose a planner-executor agent for PX4-based drones that 
decouples high-level mission planning from low-level control. A large 
language model performs single-pass task planning, while execution is 
handled through a structured ROS 2 tool-calling interface bridged to 
MAVLink. The system constructs a world model by combining modular 2D 
detectors (e.g., YOLO or vision–language models) with a pinhole depth 
projection module for 3D object localization. A constraint enforcement 
layer enforces altitude limits and horizontal geofencing, and bounded 
replanning enables recovery from execution-time action failures. We 
position our approach within three common design patterns for 
foundation-model-based robotics systems and demonstrate its 
feasibility in PX4 software-in-the-loop simulations in Gazebo. Results 
highlight improved explainability, constraint enforcement, and reduced 
LLM calls compared to tightly coupled LLM control. The code, dataset, 
videos, and other material can be found at the following link:
https://github.com/erdemuysalx/PEACE
\end{abstract}

\section{Introduction}

Recent surveys argue that integrating foundation models with UAVs 
shifts them from remotely-piloted tools towards proactive agents 
capable of natural-language interaction and contextual reasoning 
\cite{tian2025uavs,sapkota2025uavs}. Translating this promise into 
deployable systems, however, exposes three recurring bottlenecks: (i) 
keeping the foundation model in a tight perception--reasoning--action 
loop, as in ReAct-style \cite{yao2022react} agents that interleave a 
fresh LLM call with every tool invocation, inflates end-to-end latency 
and exposes the vehicle to hallucinated or ill-typed commands that can 
cause direct physical harm, motivating recent physical-safety 
benchmarks for LLM-driven UAVs \cite{tang2024defining,tian2025uavs};
(ii) compiling natural language into end-to-end visuomotor policies
yields controllers that are hard to explain, hard to constrain, and
difficult to audit against regulatory or safety requirements
\cite{tang2024defining,sapkota2025uavs}; and (iii) those same end-to-
end policies typically depend on domain-specific datasets and fine-
tuning, which limits portability across platforms and tasks, and 
amplifies the cost of development and any sim-to-real failure 
\cite{tian2025uavs,sapkota2025uavs}.

Three families dominate the current foundation-model robotics
literature, each addressing some of these bottlenecks while
inheriting others. Vision--language--navigation (VLN) systems learn
or compose policies that follow free-form navigation instructions in
previously unseen scenes 
\cite{wang2410towards,cheng2024navila,adang2025singer,lee2025cloi}. 
End-to-end vision--language--action (VLA) models such as RT-1 
\cite{brohan2022rt}, RT-2 \cite{brohan2024rt}, PaLM-E 
\cite{driess2023palm}, CognitiveDrone \cite{lykov2025cognitivedrone} 
and RaceVLA \cite{serpiva2025racevla} learn a direct mapping from 
pixels and text to control, achieving closed-loop reactivity at the 
cost of opacity and a heavy dependence on domain-specific training 
data. A third family uses foundation models as high-level planners or 
code generators that emit scripts, structured plans, or action 
primitives for a classical controller to execute, ranging from DSL-
generation systems such as TypeFly \cite{chen2023typefly} and FlockGPT 
\cite{lykov2024flockgpt} to VLM-based mission planners 
\cite{yu2025co,sautenkov2025uav}; the resulting plans are human-
readable and free of domain-specific training data, but their safety 
must be re-established for every emitted artifact.

Our work belongs to this third family. Within it, the dominant agent
paradigm is ReAct-style reasoning that interleaves a fresh LLM call
with every tool invocation, paying a model-inference cost at each
iteration and keeping the model in the fast control path. We instead
adopt the planner--executor paradigm: a single LLM pass produces a
complete, typed mission plan that the executor dispatches 
deterministically through ROS~2, with bounded replanning only on tool 
failure. This deliberate bias toward determinism and auditability 
amortizes reasoning over the whole mission, exposes the action space 
as a fixed, inspectable schema, and --- combined with an independent 
safety validation layer --- targets the latency, opacity, 
auditability, and data-dependence bottlenecks above without any task-
specific training. Our contributions are:

\begin{itemize}
  \item \emph{Planner--Executor agent} in which an LLM
  generates a complete mission plan in a single pass as a
  sequence of structured tool calls, executed without further model
  intervention. 
  \item \emph{World-grounded context building} which integrates 
  pluggable 2D detectors and a depth projection module to 
  support object-referential natural language commands grounded 
  in a live 3D world model.
  \item \emph{Safety constraint enforcement layer and recovery 
  mechanism} that enforces safety constraints such as altitude limit or
  geofence, combined with bounded replanning for execution-time action
  failures.
\end{itemize}

\section{Related Works}

We review previous work in foundation models-based robot control 
patterns across three categories: (i) vision--language--navigation 
models that enable language-conditioned navigation in unknown 
environments; (ii) vision--language--action models that enable
robots to complete tasks given natural language instructions
by the user; (iii) foundation models as high-level planners 
and code generators.

\subsection{Vision--Language--Navigation Models}
A first family focuses specifically on language-conditioned
navigation, where the robot must follow free-form instructions to
reach goals or find referents in previously unseen scenes. These
systems typically couple perception and action inside a single
learned policy, which yields strong scene-conditioned reactivity
but requires task-specific training data, ties the agent to a
fixed embodiment, and offers limited room for inserting external
safety or interpretability constraints. OpenUAV provides a platform 
and benchmark with a hierarchical multimodal LLM navigator for 
continuous 6-DOF aerial navigation \cite{wang2410towards}; NavAgent 
fuses multi-scale urban street-view information to enable embodied 
navigation in complex environments; \cite{liu2024navagent} SINGER 
trains a lightweight onboard visuomotor policy for language-guided UAV 
navigation using Gaussian-splatting simulators to reduce the sim-to-
real gap \cite{adang2025singer}. NaVILA couples a VLA model that emits 
mid-level language waypoints with low-level visual locomotion policies 
for legged robots \cite{cheng2024navila}, while CLOI-NAV targets open-
world VLN with LLM-based instruction comprehension over informative 
scene snapshots \cite{lee2025cloi}. Complementary to these, ReMEmbR 
maintains a retrieval-augmented spatio-temporal memory that an LLM 
navigator can query over long horizons \cite{anwar2025remembr}.

\subsection{Vision--Language--Action Models}
A second family of works train an end-to-end pipeline that
learns a direct mapping from pixels and language to action. The
appeal is closed-loop reactivity at control-rate frequencies, but
the resulting policies are opaque, depend on large robot-specific
datasets and fine-tuning, and are difficult to audit against
explicit safety or regulatory constraints because the action space
is implicit in the network's weights rather than exposed as a
discrete schema. RT-1 \cite{brohan2022rt} and RT-2 \cite{brohan2024rt} 
scale robotic transformers by co-fine-tuning vision--language models 
on robot and internet data; PaLM-E injects continuous sensor 
modalities into a language model for grounded planning 
\cite{driess2023palm}. For UAVs specifically, CognitiveDrone 
introduces a 7B-parameter VLA for real-time cognitive tasks 
\cite{lykov2025cognitivedrone}, and RaceVLA maps FPV video and
language to 4D velocity control for racing UAVs 
\cite{serpiva2025racevla}. Related lines include visual foundation 
models for embodied intelligence such as VC-1 \cite{majumdar2023we}, 
multi-agent coordination via retrieved skills \cite{kuroki2024multi}, 
and autonomous-driving VLAs such as EMMA \cite{hwang2024emma} and 
Alpamayo-R1 \cite{wang2025alpamayo}.

\subsection{Large Models as High-Level Planners and Code Generators for UAVs}
A third family of works uses LLMs as high-level planners that emit
scripts, structured plans, or frontier assignments, which a classical
controller then executes. Compared with the VLN and VLA approaches,
this family recovers interpretability and removes the dependence
on robot-specific training data, but most existing instances either
emit free-form code or a custom DSL that must be re-validated for
every mission, or follow a ReAct-style loop that re-invokes the LLM
at every step, and reintroduces inference latency on the control
path. TypeFly generates flight plans in a domain-specific language 
(MiniSpec) to reduce LLM inference latency \cite{chen2023typefly}. 
FlockGPT translates natural-language instructions into signed distance 
function targets that guide UAV swarms \cite{lykov2024flockgpt}.
Co-NavGPT uses a vision--language model as a global planner with
frontier-based visual prompting for multi-robot target search
\cite{yu2025co}. LLM-based mission planners have also been proposed for
extreme-environment exploration \cite{sautenkov2025uav}, and surveys 
and demonstrators in swarm robotics cover language-conditioned code
generation and pattern formation via multi-agent reinforcement
learning \cite{strobel2024llm2swarm,liu2024language}. In our work, we 
forgo the closed-loop reactivity and fine-grained scene conditioning 
of a learned VLN/VLA policy in exchange for explainability, 
auditability, and low control-path latency: an off-the-shelf LLM is 
exposed as a single-pass planner that emits a typed
\texttt{MissionPlan} rather than free-form code or a custom DSL, the 
executor binds each tool directly to a MAVLink via ROS~2, and an 
independent safety validation service validates every motion command 
before actuation. The action space is therefore fixed, inspectable, 
and decoupled from the model's weights, which we view as the right 
point on the expressiveness--verifiability frontier for safety-
critical UAV deployment.

\section{Approach}

We develop an agent to generate mission plans for an open world
UAV flight via natural language. World model representations are 
generated using a perception backend of choice: a pretrained object 
detection model or a vision--language--model. Robot states are 
aggregated from real-time sensor readings. We fuse the world model and the robot state histories into a system prompt to be used with the large--language--model to generate the mission plan.

\subsection{System Overview}
\begin{figure*}[t]
  \centering
  \includegraphics[width=1\linewidth]{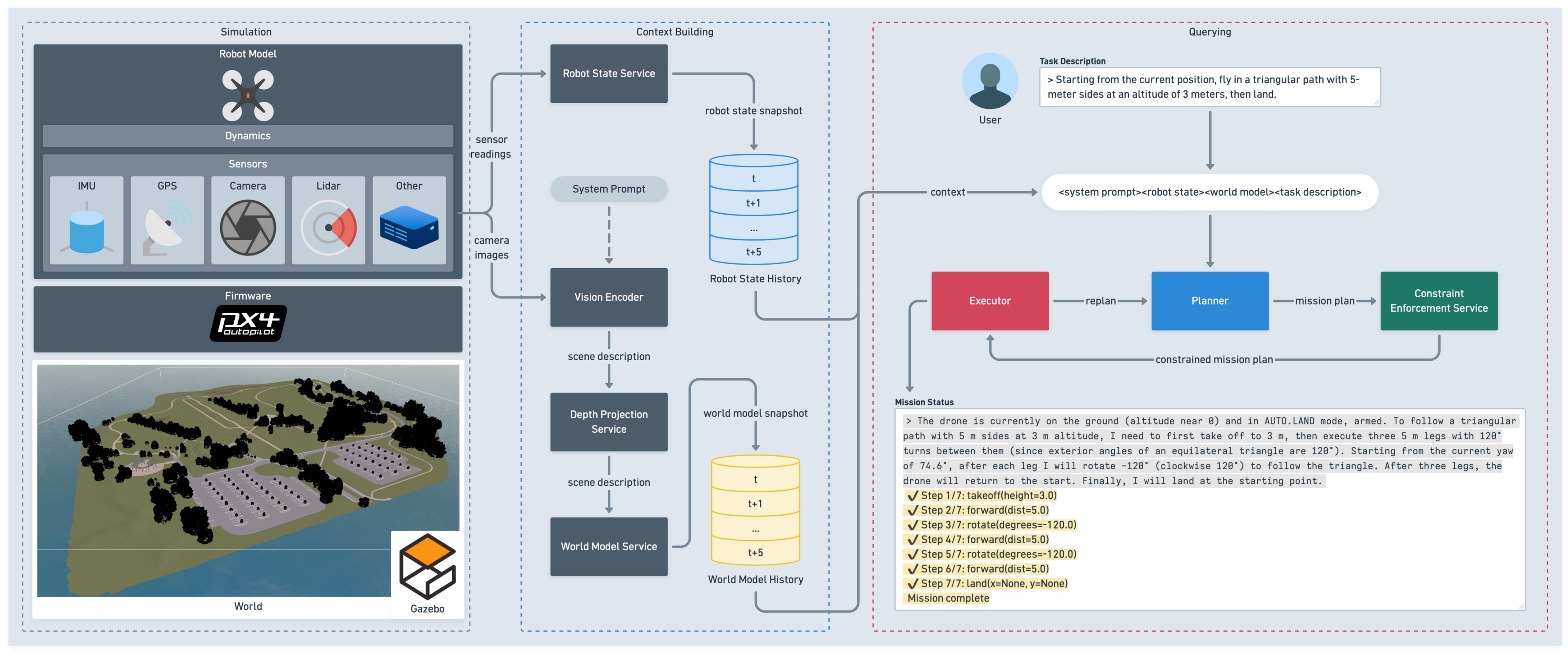}
  \caption{System architecture overview. The system comprises three modules. (Left) The simulation environment runs a drone model with IMU, GPS, depth camera, and LiDAR sensors in Gazebo, with PX4 Autopilot handling low-level flight control. (Center) The context-building pipeline aggregates sensor readings into a Robot State Service and passes camera images through a Vision Encoder and Depth Projection Service to maintain a timestamped Robot State History and World Model History. (Right) At query time, the system prompt, robot state, world model, and user task description are composed into a single context for the Planner, which generates a mission plan validated by the Constraint Enforcement Service before being dispatched to the Executor. Execution failures trigger bounded replanning via a feedback loop to the Planner.}
  \label{fig:overview}
\end{figure*}

The agent comprises four conceptual components: a
\emph{planner--executor} that turns natural-language queries into
mission plans and dispatches them, a \emph{perception} module that
produces detections and grounds them in 3D, a \emph{tool set} that
binds a fixed action space, perception, and life-cycle operations to 
the flight controller, and a set of shared services that aggregate the 
world model and robot state and validate every motion command against 
safety constraints. Figure~\ref{fig:overview} sketches the data flow.

\subsection{Planner--Executor Architecture}
A single LLM call converts the user query, robot state, and world
model into a structured mission plan: a reasoning string followed by an
ordered list of tool calls. The executor dispatches the tool calls
sequentially and re-invokes the LLM only on tool failure, in which
case a bounded replan is issued. Removing the LLM from the fast
control path eliminates an entire class of latency-induced and
hallucination-induced failure modes reported in reactive
loop-in-the-LLM systems \cite{tang2024defining,tian2025uavs}.

\subsection{Perception Backends and 3D Grounding}
Perception is abstracted behind a detector interface with two
interchangeable backends: a fast 2D object detector and a
vision--language model generating structured detections. A
pinhole-camera depth projector lifts detections to ENU-frame 3D
positions using the synchronized depth image and camera intrinsics,
and exposes an explicit depth-validity flag so the planner can
distinguish confident localizations from uncertain ones. The
abstraction keeps perception pluggable, in the spirit of recent
pretrained-representation \cite{majumdar2023we} and VLN
\cite{adang2025singer,lee2025cloi} work, without committing to any specific VLA backbone.

\subsection{World Model and Robot-State Aggregation}
A world-model service aggregates the detection stream into a bounded
history of timestamped scene snapshots, and a parallel robot-state
service aggregates odometry, flight mode, and arming state. Each
planner invocation reads a consistent snapshot from both services,
which is the property required to ground the system prompt against
the world as it is at planning time \cite{anwar2025remembr}.

\subsection{Prompt Construction and Context Assembly}
Each planner invocation is a single LLM call with two messages, 
decomposed by their lifetime. The \emph{system prompt} is
static and carries the invariant knowledge: the tool set, the 
mission rules, and the output schema. The \emph{user prompt} is
assembled at query time and carries only the volatile states:
the natural-language task description, the current robot state, and the
current world model, with a short history of past states for temporal 
reasoning. On replan, the user message is rebuilt from a fresh 
snapshot so that subsequent planning steps see the world as it is at 
the moment of replanning rather than as it was when the mission began.

\subsection{Constraint Enforcement}
A dedicated safety layer validates every motion command before it
reaches the autopilot, enforcing two geometric constraints: an
altitude band and a horizontal geofence around the takeoff point.
Out-of-band altitude targets are clamped; geofence breaches abort
the offending tool calls and surfaces a structured failure to the
replanner. Because the check is purely geometric, it is verifiable
independently of the language model, and provides the construction
required by the avoid-collision and regulatory-compliance axes of
recent physical-safety benchmarks for LLM-controlled UAVs
\cite{tang2024defining}.

\section{Experiments}

We report preliminary experiments of the proposed agent. The aim
is not a quantitative benchmark, but to characterize the agent's
behavior over a structured set of natural-language prompts and to
surface the recurring failure modes that constrain the current
design. 

\subsection{Experimental Procedure}
We compiled a set of 108 natural-language prompts spanning eight
task categories, aligned with the capabilities exposed by the tool
set: \emph{takeoff and landing} (12), \emph{relative motion} (12),
\emph{altitude and attitude control} (12), \emph{waypoint navigation} 
(13), \emph{geometric path following} (13), \emph{object search and
localization} (12), \emph{open-vocabulary navigation} (13),
\emph{composite multi-step missions} (13), and \emph{safety violation} 
(8). Each prompt carries a stable identifier and is phrased as a 
single, free-form natural-language instruction; representative 
examples are ``Fly to position $x=3,\,y=4,\,z=3$'' for waypoint 
navigation, rest can be obtained via 
\footnote{https://huggingface.co/datasets/erdemuysalx/uav-mission-
prompts}. Object-referential prompts are restricted to the vocabulary 
\emph{\{person, car, bench, traffic light, fire hydrant\}}, due to the 
availability of assets in the simulation.

\subsection{Preliminary Results}
The mission logging makes each mission auditable offline. For the 
prompt ``Take off to 5 meters then move forward 3 meters'', for 
example, the planner emits a single \texttt{MissionPlan} comprising
\texttt{takeoff(5)} followed by \texttt{forward(3)}; the executor
dispatches the two calls in order, and the recorded reasoning-step
results, and the final robot state together describe what the agent
attempted, what it did, and where it ended up.

Across categories, four behavioral patterns are visible. Metric
tasks (waypoint, relative motion, altitude control) admit the most
reliable plans, since the parameters in the prompt map one-to-one
onto a tool argument. Geometric paths are correctly decomposed into
ordered waypoints when the shape and edge length are stated 
explicitly, and the trajectory is resolved at planning time, 
consistent with the open-loop design. Perception-conditioned 
categories add a dependence on detector recall and on the
depth-validity flag exposed by the world model. In composite missions, 
steps depend on each other, so an undetected failure in one step can 
cascade into mission failure.

\subsection{Failure Modes}
The preliminary results exhibit three recurring failure modes. (i)
\emph{Planner-level errors} arise when the LLM emits a plan that
violates the typed schema or a mission rule encoded in the system
prompt; such plans are caught at validation and either fail
terminally or trigger a bounded replan. (ii) \emph{Perception-level
uncertainty} arises when the requested object class is absent from
the detection set or marked as uncertain by the depth projector; the
planner avoids acting on uncertain detections, but extended runs of
uncertain observations, collapse object search and navigation to
degenerate plans. (iii)\emph{ Safety-validation aborts}, where an 
action is rejected because it would breach the altitude limit or the 
horizontal geofence; these are by design but propagate to the mission
failure.

\section{Implications and Future Directions}

We have presented an agent in which mission planning is decoupled
from control, and in which every action that reaches the low-level 
controller is structured and safety constrained by an independent 
service. This design point is distinct from code-generating planners 
\cite{chen2023typefly,lykov2024flockgpt}, which emit executable 
artifacts for every mission  and thus expand the action space at run 
time, and from end-to-end vision--language--action policies
\cite{lykov2025cognitivedrone,serpiva2025racevla,brohan2022rt,brohan2024rt}, which embed the action space in the network weights and offer no
inspectable trace of what the model decided.

The separation between the language model and the control path has
three practical implications. First, mission plans become
auditable: the planner's reasoning, the structured mission plan, and 
the per-step results are recoverable from the mission log, which makes
offline analysis of failure modes possible. Second, an off-the-shelf 
LLM is sufficient for the planner; in-context-learning composition of 
natural-language tasks over the tool set requires no domain-specific 
fine-tuning, in contrast to domain-specific models 
\cite{brohan2022rt,brohan2024rt,lykov2025cognitivedrone,serpiva2025racevla}.
Third, safety constraints are verified outside the planner, so 
violations are caught even when the planner emits a plan that is 
otherwise well-formed. Moreover, the adoption of a planner–executor
paradigm reduces LLM invocations to a single call, which is
substantially lower than ReAct-based agents that typically require an 
LLM invocation after each tool interaction.

Three limitations are visible in the failure modes of Section IV.
Perception-level uncertainty manifests because the world model is the 
only feedback channel between the scene and the planner, so a sequence 
of degenerate snapshots can lead to a degenerate plan that no recovery 
mechanism below the granularity of a tool failure can correct. 
Execution-level state-machine interactions, particularly between off-
board control and arming, indicate that idempotency at the tool layer 
is required: each tool should re-establish its preconditions 
internally rather than depend on the planner to interleave them.

Two directions follow from the design above. The first is a
sim-to-real deployment on physical UAV hardware. The second concerns 
dynamic scenes whose state changes within a mission: the current open-
loop planning assumption is brittle in such settings, since a target
object that moves between planning and execution, or an obstacle
that appears mid-mission, is only handled at the granularity of a
tool failure. The accumulated mission traces, consisting of plans 
paired with perception snapshots and outcomes, are suitable for 
distilling a compact vision--language--action model that can complement
the language-conditioned planner with a reactive, visually-grounded
channel for dynamic environments.

\section*{Acknowledgment}

We thank the EU Commission and the State Secretariat 
for Education, Research, and Innovation (SERI) for funding the 
\href{https://www.innoguard.eu/index.html}{project InnoGuard}, Marie
Sktodowska-Curie Actions Doctoral Networks (HORIZON-MSCA-2023-DN), 
the Swiss National Science Foundation (SNSF) for funding the  "SwarmOps" project (No. 200021\_219732). 

\bibliographystyle{IEEEtran} 
\bibliography{references}

@inproceedings{yao2022react,
  title={React: Synergizing reasoning and acting in language models},
  author={Yao, Shunyu and Zhao, Jeffrey and Yu, Dian and Du, Nan and Shafran, Izhak and Narasimhan, Karthik R and Cao, Yuan},
  booktitle={The eleventh international conference on learning representations},
  year={2022}
}

@article{strobel2024llm2swarm,
  title={LLM2Swarm: robot swarms that responsively reason, plan, and collaborate through LLMs},
  author={Strobel, Volker and Dorigo, Marco and Fritz, Mario},
  journal={arXiv preprint arXiv:2410.11387},
  year={2024}
}

@article{yu2025co,
  title={Co-NavGPT: Multirobot Cooperative Visual Semantic Navigation Using Vision Language Models},
  author={Yu, Bangguo and Yuan, Qihao and Li, Kailai and Kasaei, Hamidreza and Cao, Ming},
  journal={IEEE Robotics and Automation Letters},
  volume={11},
  number={2},
  pages={2122--2129},
  year={2025},
  publisher={IEEE}
}

@inproceedings{anwar2025remembr,
  title={Remembr: Building and reasoning over long-horizon spatio-temporal memory for robot navigation},
  author={Anwar, Abrar and Welsh, John and Biswas, Joydeep and Pouya, Soha and Chang, Yan},
  booktitle={2025 IEEE International Conference on Robotics and Automation (ICRA)},
  pages={2838--2845},
  year={2025},
  organization={IEEE}
}

@article{cheng2024navila,
  title={Navila: Legged robot vision-language-action model for navigation},
  author={Cheng, An-Chieh and Ji, Yandong and Yang, Zhaojing and Gongye, Zaitian and Zou, Xueyan and Kautz, Jan and B{\i}y{\i}k, Erdem and Yin, Hongxu and Liu, Sifei and Wang, Xiaolong},
  journal={arXiv preprint arXiv:2412.04453},
  year={2024}
}

@inproceedings{lee2025cloi,
  title={CLOI-NAV: Open-World Vision-and-Language Navigation via Complex, Long-horizon Ordered Instructions},
  author={Lee, Minho and Park, Jaeil and Jeong, Jinyong and Cho, Younggun},
  booktitle={IROS 2025 Workshop: Open World Navigation in Human-centric Environments},
  year={2025}
}

@inproceedings{kuroki2024multi,
  title={Multi-agent behavior retrieval: Retrieval-augmented policy training for cooperative push manipulation by mobile robots},
  author={Kuroki, So and Nishimura, Mai and Kozuno, Tadashi},
  booktitle={2024 IEEE/RSJ International Conference on Intelligent Robots and Systems (IROS)},
  pages={12671--12678},
  year={2024},
  organization={IEEE}
}

@article{sapkota2025uavs,
  title={UAVs meet agentic AI: A multidomain survey of autonomous aerial intelligence and agentic UAVs},
  author={Sapkota, Ranjan and Roumeliotis, Konstantinos I and Karkee, Manoj},
  journal={arXiv preprint arXiv:2506.08045},
  year={2025}
}

@article{tian2025uavs,
  title={UAVs meet LLMs: Overviews and perspectives towards agentic low-altitude mobility},
  author={Tian, Yonglin and Lin, Fei and Li, Yiduo and Zhang, Tengchao and Zhang, Qiyao and Fu, Xuan and Huang, Jun and Dai, Xingyuan and Wang, Yutong and Tian, Chunwei and others},
  journal={Information Fusion},
  volume={122},
  pages={103158},
  year={2025},
  publisher={Elsevier}
}

@article{chen2023typefly,
  title={Typefly: Flying drones with large language model},
  author={Chen, Guojun and Yu, Xiaojing and Ling, Neiwen and Zhong, Lin},
  journal={arXiv preprint arXiv:2312.14950},
  year={2023}
}

@article{wang2410towards,
  title={Towards realistic uav vision-language navigation: Platform, benchmark, and methodology. arXiv 2024},
  author={Wang, X and Yang, D and Wang, Z and Kwan, H and Chen, J and Wu, W and Li, H and Liao, Y and Liu, S},
  journal={arXiv preprint arXiv:2410.07087},
  year={2024}
}

@article{liu2024navagent,
  title={Navagent: Multi-scale urban street view fusion for uav embodied vision-and-language navigation},
  author={Liu, Youzhi and Yao, Fanglong and Yue, Yuanchang and Xu, Guangluan and Sun, Xian and Fu, Kun},
  journal={arXiv preprint arXiv:2411.08579},
  year={2024}
}

@inproceedings{lykov2024flockgpt,
  title={FlockGPT: Guiding UAV flocking with linguistic orchestration},
  author={Lykov, Artem and Karaf, Sausar and Martynov, Mikhail and Serpiva, Valerii and Fedoseev, Aleksey and Konenkov, Mikhail and Tsetserukou, Dzmitry},
  booktitle={2024 IEEE International Symposium on Mixed and Augmented Reality Adjunct (ISMAR-Adjunct)},
  pages={485--488},
  year={2024},
  organization={IEEE}
}

@article{adang2025singer,
  title={SINGER: An onboard generalist vision-language navigation policy for drones},
  author={Adang, Maximilian and Low, JunEn and Shorinwa, Ola and Schwager, Mac},
  journal={arXiv preprint arXiv:2509.18610},
  year={2025}
}

@article{tang2024defining,
  title={Defining and evaluating physical safety for large language models},
  author={Tang, Yung-Chen and Chen, Pin-Yu and Ho, Tsung-Yi},
  journal={arXiv preprint arXiv:2411.02317},
  year={2024}
}

@article{serpiva2025racevla,
  title={RaceVLA: VLA-based racing drone navigation with human-like behaviour},
  author={Serpiva, Valerii and Lykov, Artem and Myshlyaev, Artyom and Khan, Muhammad Haris and Abdulkarim, Ali Alridha and Sautenkov, Oleg and Tsetserukou, Dzmitry},
  journal={arXiv preprint arXiv:2503.02572},
  year={2025}
}

@article{lykov2025cognitivedrone,
  title={Cognitivedrone: A vla model and evaluation benchmark for real-time cognitive task solving and reasoning in uavs},
  author={Lykov, Artem and Serpiva, Valerii and Khan, Muhammad Haris and Sautenkov, Oleg and Myshlyaev, Artyom and Tadevosyan, Grik and Yaqoot, Yasheerah and Tsetserukou, Dzmitry},
  journal={arXiv preprint arXiv:2503.01378},
  year={2025}
}

@inproceedings{liu2024language,
  title={Language-guided pattern formation for swarm robotics with multi-agent reinforcement learning},
  author={Liu, Hsu-Shen and Kuroki, So and Kozuno, Tadashi and Sun, Wei-Fang and Lee, Chun-Yi},
  booktitle={2024 IEEE/RSJ International Conference on Intelligent Robots and Systems (IROS)},
  pages={8998--9005},
  year={2024},
  organization={IEEE}
}

@article{driess2023palm,
  title={Palm-e: An embodied multimodal language model},
  author={Driess, Danny and Xia, Fei and Sajjadi, Mehdi SM and Lynch, Corey and Chowdhery, Aakanksha and Ichter, Brian and Wahid, Ayzaan and Tompson, Jonathan and Vuong, Quan and Yu, Tianhe and others},
  journal={arXiv preprint arXiv:2303.03378},
  year={2023}
}

@article{brohan2022rt,
  title={Rt-1: Robotics transformer for real-world control at scale},
  author={Brohan, Anthony and Brown, Noah and Carbajal, Justice and Chebotar, Yevgen and Dabis, Joseph and Finn, Chelsea and Gopalakrishnan, Keerthana and Hausman, Karol and Herzog, Alex and Hsu, Jasmine and others},
  journal={arXiv preprint arXiv:2212.06817},
  year={2022}
}

@article{brohan2024rt,
  title={Rt-2: Vision-language-action models transfer web knowledge to robotic control, 2023},
  author={Brohan, Anthony and Brown, Noah and Carbajal, Justice and Chebotar, Yevgen and Chen, Xi and Choromanski, Krzysztof and Ding, Tianli and Driess, Danny and Dubey, Avinava and Finn, Chelsea and others},
  journal={URL https://arxiv.org/abs/2307.15818},
  volume={1},
  pages={2},
  year={2024}
}

@article{majumdar2023we,
  title={Where are we in the search for an artificial visual cortex for embodied intelligence?},
  author={Majumdar, Arjun and Yadav, Karmesh and Arnaud, Sergio and Ma, Jason and Chen, Claire and Silwal, Sneha and Jain, Aryan and Berges, Vincent-Pierre and Wu, Tingfan and Vakil, Jay and others},
  journal={Advances in Neural Information Processing Systems},
  volume={36},
  pages={655--677},
  year={2023}
}

@article{hwang2024emma,
  title={Emma: End-to-end multimodal model for autonomous driving},
  author={Hwang, Jyh-Jing and Xu, Runsheng and Lin, Hubert and Hung, Wei-Chih and Ji, Jingwei and Choi, Kristy and Huang, Di and He, Tong and Covington, Paul and Sapp, Benjamin and others},
  journal={arXiv preprint arXiv:2410.23262},
  year={2024}
}

@article{wang2025alpamayo,
  title={Alpamayo-r1: Bridging reasoning and action prediction for generalizable autonomous driving in the long tail},
  author={Wang, Yan and Luo, Wenjie and Bai, Junjie and Cao, Yulong and Che, Tong and Chen, Ke and Chen, Yuxiao and Diamond, Jenna and Ding, Yifan and Ding, Wenhao and others},
  journal={arXiv preprint arXiv:2511.00088},
  year={2025}
}

@article{sautenkov2025uav,
  title={Uav-codeagents: Scalable uav mission planning via multi-agent react and vision-language reasoning},
  author={Sautenkov, Oleg and Yaqoot, Yasheerah and Mustafa, Muhammad Ahsan and Batool, Faryal and Sam, Jeffrin and Lykov, Artem and Wen, Chih-Yung and Tsetserukou, Dzmitry},
  journal={arXiv preprint arXiv:2505.07236},
  year={2025}
}

\end{document}